\documentclass{article}

\usepackage{times}
\usepackage{graphicx} 
\usepackage{setspace}    
\usepackage[intlimits]{amsmath} 
\usepackage{amsxtra}     
\usepackage{amsmath,amsfonts,amssymb}
\usepackage{natbib}
\usepackage{algorithm}
\usepackage{algorithmic}
\usepackage{hyperref}
\usepackage{dblfloatfix}
\usepackage{booktabs}

\usepackage{url}

\newcommand{\R}{\mathbb{R}}
\DeclareMathOperator*{\argmax}{arg\,max}
\DeclareMathOperator*{\argmin}{arg\,min}

\usepackage[accepted]{icml2016}

\newcommand{\cutsectionup}{\vspace*{-0.08in}}

\begin{document} 
\thispagestyle{plain}
\pagestyle{plain}
\twocolumn[
\icmltitle{Text-to-Image-to-Text Translation using Cycle Consistent Adversarial Networks}

\icmlauthor{Satya Krishna Gorti, Jeremy Ma}{ \{satyag, junweima\}@cs.toronto.edu}
\icmladdress{University of Toronto}

\vskip 0.3in
]

\begin{abstract} 
Text-to-Image translation has been an active area of research in the recent past. The ability for a network to learn the
meaning of a sentence and generate an accurate image that depicts the sentence shows ability of the model to think more like humans. Popular methods on text to image translation make use of Generative Adversarial Networks (GANs) to generate high quality images based on text input, but the generated images don't always reflect the meaning of the sentence given to the model as input. We address this issue by using a captioning network to caption on generated images and exploit the distance between ground truth captions and generated captions to improve the network further. We show extensive comparisons between our method and existing methods.\footnote{Authors contributed equally}

\vspace*{-0.15in}
\end{abstract} 

\cutsectionup
\section{Introduction}

Text-to-Image synthesis is a challenging problem that has a lot of room for improvement considering the current state-of-the-art results. Synthesized images from existing methods give a rough sketch of the described image but fail to capture the true essence of what the text describes. The recent success of Generative Adversarial Networks (GANs) \cite{goodfellow2014generative} indicate that they are a good candidate for the choice of architecture to approach this problem.

However, the very nature of this problem is such that a piece of text can map to multiple valid images. The lack of such a direct one-to-one mapping means that traditional conditional GANs \cite{mirza2014conditional} cannot be used directly. We draw our inspiration from the recent works of image-to-image translation \cite{liu2017unsupervised}\cite{zhu2017unpaired} where cycle consistent GANs have been trained and achieved very impressive results.

Reproducing results from recent work \cite{reed2016generative} \cite{zhang2017stackgan}, we observed that a lot of effort has been put into making the network produce high quality images. But we found that many images generated  are still not accurate to the ground truth text descriptions given as input to the network.

We address this problem by having a framework similar to CycleGAN \cite{zhu2017unpaired}. We first break the image synthesis network into two stages similar to a StackGAN architecture \cite{zhang2017stackgan}. This consists of generating a low resolution image in the first stage and feeding this as input to the next stage's generator. The next stage refines the low resolution input further and generates higher quality image with 128x128 resolution. 

\begin{figure*}
\begin{center}
	\includegraphics[height=5cm,width=10cm]{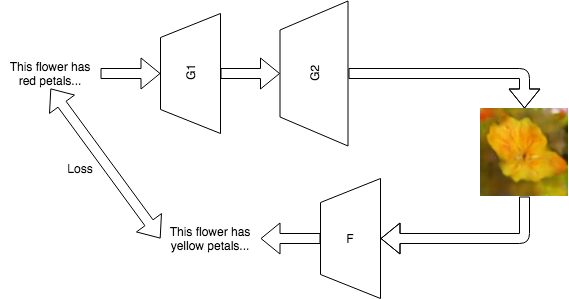}
    \caption{High level diagram of the design. G1, G2 are the stage 1 and stage 2 generator. F is the caption generator. We take the difference between the original caption and the real caption as the loss.}
    \label{fig:overallarc}
\end{center}
\end{figure*}

We then implement an image captioning GAN similar to \cite{DBLP:journals/corr/DaiLUF17}. This network generates high quality captions based on images. We finally observe the difference in the ground truth captions and captions generated by our image captioning network. This provides a good signal for optimizing the image synthesis network further to generate good images that represent the text descriptions well. Figure \ref{fig:overallarc} shows the high level design of our system.


Therefore, we have two generators $G$ (broken down further into two stages $G1$ and $G2$) and $F$. We train a mapping $G: T_{emb} \mapsto Y$ and inverse mapping $F: Y \mapsto T$ in a cycle consistent manner, where $T_{emb}$ is a fixed length embedding for the text that describes an image generated by Skip-Thought Vector network \cite{kiros2015skip} and $T$ represents the caption. The generators $G$ and $F$ have their corresponding discriminator $D$ (broken down into $D1$ and $D2$) and $E$.

We show that the results generated by training the network in a cycle consistent manner produces more relevant images based on ground truth text descriptions. We also tabulate the inception scores of the images generated by our network.

\section{Related work}

Generative Adversarial Networks (GANs) have achieved impressive results in problems such as image generation \cite{DBLP:journals/corr/RadfordMC15}. Conditional GANs introduced in \cite{mirza2014conditional} build on top of GANs by learning to approximate the distribution of data by conditioning on an input.

In the recent past there have been attempts on text to image synthesis using conditional GANs such as \cite{reed2016generative}\cite{dash2017tac}\cite{zhang2017stackgan}\cite{DBLP:journals/corr/abs-1710-10916}. We can see promising results in Reed et al. \cite{reed2016generative} by conditioning the GAN on text descriptions instead of class labels. Their follow-up work \cite{NIPS2016_6111} added additional annotations on object part locations and generated successfully images of resolution 128x128. \cite{dash2017tac} additionally conditions its generative process with both text and class information and has produced superior results compared to \cite{reed2016generative}. 

Zhang et al. \cite{zhang2017stackgan} used a similar approach but break the process of generation down into a two stage process. Stage-1 network consists of a conditional GAN, where the generator produces low-resolution 64x64 images based on text description. The Stage-1 discriminator distinguishes between real and fake 64x4 images. Stage-2 takes Stage-1 generator's result as input and generates high resolution photo-realistic images of resolution 256x256. The Stage-2 discriminator as expected learns to distinguish between real and fake 256x256 images.

Recently, some attempts have also been made to use a conditional GAN to to generate realistic captions based on images, most notably by Dai et al. \cite{DBLP:journals/corr/DaiLUF17}. Here, the generator network is made to produce high quality captions on images and an evaluator network is made to assess the quality of captions based on the visual content.

We also need to have an effective fixed length representation for the text description of images. Popular choice of text embeddings for text-to-image translation in recent literature are: Skip Thought Vectors introduced by Kiros et al. \cite{kiros2015skip} and embeddings produced by Char-CNN-RNN model, a hybrid character-level ConvNet with a recurrent neural network introduced by Reed et al. \cite{DBLP:journals/corr/ReedASL16}. We use Skip-Thought Vectors to represent text.

Cycle consistent GANs have showed excellent results for multimodal learning problems, which lack a direct one-to-one correspondence with input and output and allows the network to learn many mappings at the same time as shown in \cite{zhu2017unpaired}\cite{liu2017unsupervised}\cite{zhou2016learning}. 
Zhu et al. \cite{zhu2017unpaired} learn a mapping $G: X \to Y$ such that the network can translate images from domain $X$ to images similar the distribution of domain $Y$ in absence of paired images from these distributions. Since it is difficult to learn this under-constrained mapping directly, \cite{zhu2017unpaired} introduce a forward cycle consistency loss $F(G(X)) \approx X$ and backward cycle consistency loss $G(F(Y)) \approx Y$ where F is a mapping represented by: $F : Y \to X$.
To the best of our knowledge, CycleGAN still hasn't been used for text-to-image generation, which is a similar multimodal learning problem.

\section{Method}
\subsection{Text to Image Translation}

We build the conditional GAN by using conditioning the Skip-Thought text embeddings \cite{kiros2015skip} as input to both the generator and the discriminator.

Since we wanted to produce high resolution photo-realistic images, we model our GAN with an architecture that is similar to StackGAN \cite{zhang2017stackgan}. In this architecture, the GAN is separated out into two stages, Stage-1 and Stage-2. Each stage consists of a generator, discriminator pair $G_1$ and $D_1$ and $G_2$ and $D_2$.

\subsubsection{Network Architecture}
\begin{figure*}
\begin{center}
	\includegraphics[height=10cm,width=13cm]{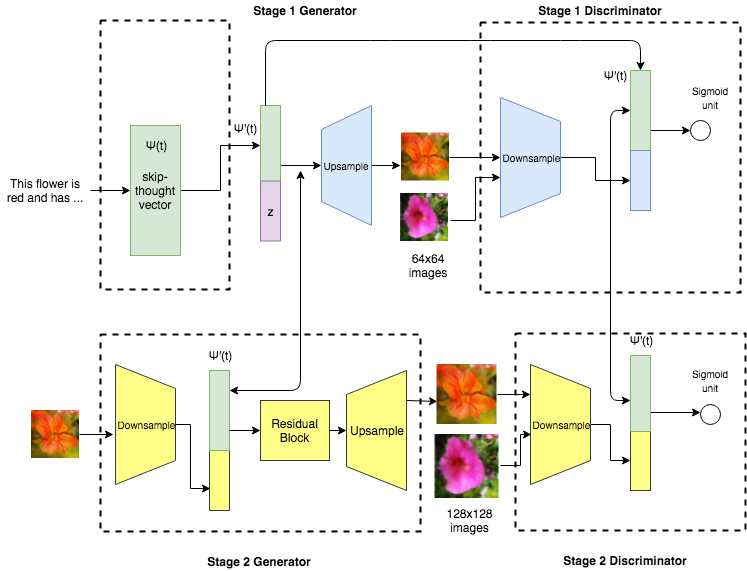}
    \caption{Text to Image GAN network. The text embedding and noise is given as input to the first stage. The output of the first stage is given as input to the next stage that produces higher resolution images. Generators from each stage have corresponding discriminators}
    \label{texttoimagearc}
\end{center}
\end{figure*}
\begin{figure*}
\begin{center}
	\includegraphics[height=9.5cm,width=12cm]{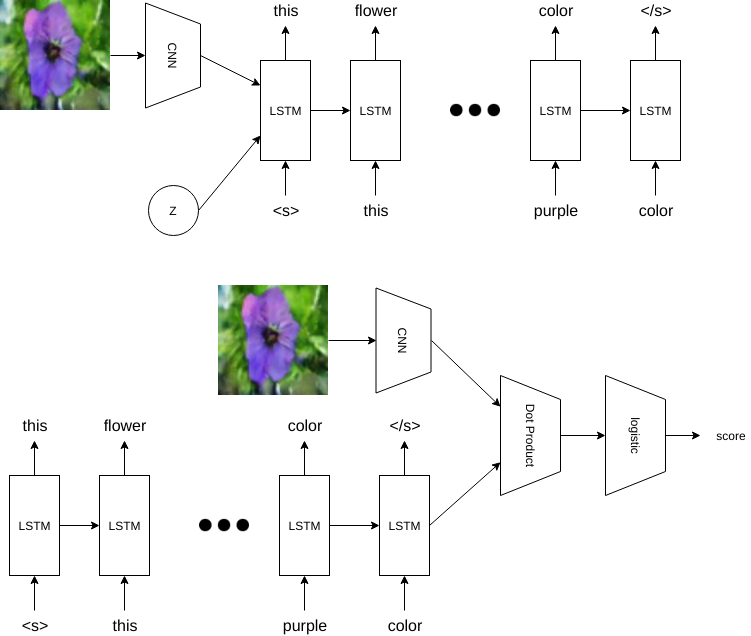}
    \caption{Caption GAN network. The top row shows the caption generator where the LSTM takes CNN features and noise Z as input and outputs captions. The bottom row shows the discriminator that performs a dot product on the CNN features of the image and the LSTM output.}
    \label{imagetotextarc}
\end{center}
\end{figure*}
We condition $G_1$ on the skip thought text embedding $\psi(t) \in \R^T$ and generate a synthetic image $I_{1}$ with a low resolution of 64x64. The stage 1 discriminator $D_1$ is conditioned on the same text embedding $\psi(t)$ and is trained to classify between real and synthetic images of resolution 64x64. The Stage-2 generator $G_1$ takes $I_1$ as input along with the embedding $\psi(t)$ and generates higher resolution 128x128 image $I_2$. The corresponding takes $I_2$ as input the rest of its architecture is the similar to $D_1$

Hence generators $G_1$ and $G_2$ are denoted by $G_1: \R^{Z}\times\R^{T} \to \R^{64\times64}$ and $G_2: \R^{Z}\times\R^{T} \to \R^{128\times128}$. Discriminators $D_1$ and $D_2$ are denoted by  $D_1: \R^{64\times64}\times\R^{T} \to \{0, 1\}$ and $\R^{128\times128}\times\R^{T} \to \{0, 1\}$. Here we sample from random noise prior $z \in \R^Z \thicksim  \mathcal{N}(0,1)$. 

Our text embedding $\psi(t)$ is 2400 dimension vector. We compress this to a small dimension (128) using a fully connected layer followed by a non linear activation (Leaky-ReLU) and concatenate it with the noise vector $z$. We feed forward this vector as input to both generators $G_1$ and $G_2$.

The generator $G_1$ consists of a series of upsampling blocks. These blocks consist of nearest-neighbor upsampling followed by a $3\times3$ stride 1 convolution to project the input to an 3x64x64 image $I_1$.

The discriminator $D_1$ consists of a series of downsampling blocks that project the input to a dimension of 512x4x4. We then concatenate this with the 128 dimensional compressed embedding and use a sigmoid layer to produce an output between 0 and 1.

The generator $G_2$ consists of a series of downsampling blocks that first project the 3x64x64 input image to a dimension of 512x16x16. We then concatenate the 128 dimensional embedding. This is sent as input a series of residual blocks followed by a series of upsampling blocks similar to those in $G_1$.

\subsubsection{Stage-1 GAN}

We train the Stage-1 GAN by maximizing $\mathcal{L}_{D1}$ and minimizing $\mathcal{L}_{G1}$ given in equations \ref{eq:stage1d} and \ref{eq:stage1g}.
\begin{equation} \label{eq:stage1d}
\begin{aligned}
\mathcal{L}_{D1} = \mathbb{E}_{(I_1, \psi(t)) \thicksim p_{data}}[\log D_1(I_1, \psi(t)] +\\
\mathbb{E}_{I_1 \thicksim G_1, \psi(t) \thicksim p_{data}}[\log (1 - D_1(I_1, \psi(t))]
\end{aligned}
\end{equation}
\begin{equation} \label{eq:stage1g}
\begin{aligned}
\mathcal{L}_{G1} = \mathbb{E}_{z \thicksim \mathcal{N}(0,1), \psi(t) \thicksim p_{data}}[\log (1 - D_1(G_1(z, \psi(t)))]
\end{aligned}
\end{equation}

\subsubsection{Manifold Interpolation Loss}
\begin{figure*}[t!]
\begin{center}
	\includegraphics[width=\textwidth]{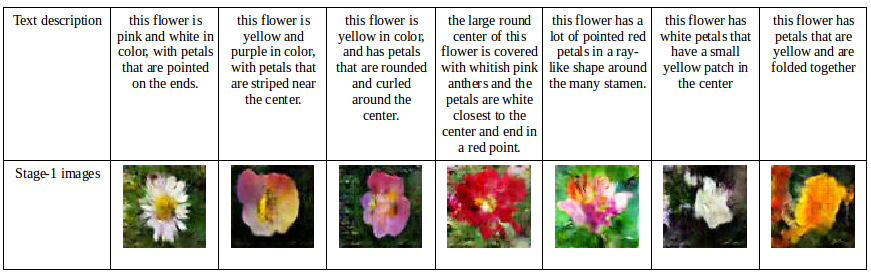}
    \caption{Stage 1 GAN results which are 64x64 images}
    \label{fig:stage1}
\end{center}
\end{figure*}
\begin{figure*}[t!]
\begin{center}
	\includegraphics[width=\textwidth]{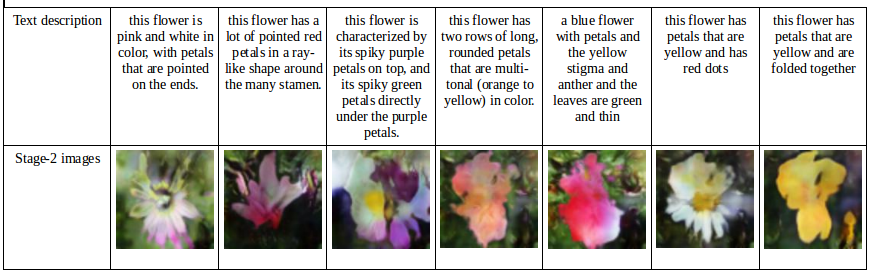}
    \caption{Stage 2 GAN results which are 128x128 images}
    \label{fig:stage2}
\end{center}
\end{figure*}
Reed et al. \cite{reed2016generative} introduced an additional interpolation loss. This loss term is based on using embeddings obtained by interpolating between embeddings of the training set. Since these interpolated embeddings tend to be near the data manifold in deep networks, they could correspond to a valid image from the image distribution. Based on this assumption we should be able to fool the discriminator by using an interpolated text embedding given by $\beta \psi(t1) + (1-\beta \psi(t2)$, where $\beta$ is a constant which we set to 0.5. In practice we found using this to produce superior results. Hence we have to minimize an additional interpolation loss term $\mathcal{L}_{INT}$ given in equation \ref{eq:intloss}
\begin{equation} \label{eq:intloss}
\begin{aligned}
\mathcal{L}_{INT} = \mathbb{E}_{(\psi(t1), \psi(t2)) \thicksim p_{data}}[\log (1 - \\
D_1(G(z, \beta \psi(t1) + (1-\beta \psi(t2)]
\end{aligned}
\end{equation}

Therefore we have to total loss $\mathcal{L}_{G1-Total}$ for training the generator in Stage-1 given by equation \ref{eq:stage1total}.
\begin{equation} \label{eq:stage1total}
\begin{aligned}
\mathcal{L}_{G1-Total} = \mathcal{L}_{G1} + \lambda \mathcal{L}_{INT}
\end{aligned}
\end{equation}

Where $\lambda$ is hyperparameter that determines the importance given to $\mathcal{L}_{INT}$ loss. We use $\lambda = 0.5$.

\subsubsection{Stage-2 GAN}

The Stage-2 GAN improves the results further by taking the low-resolution images generated in Stage-1 and the actual text embedding and further learns features that were ignored by Stage-1 generator.

We train the Stage-2 GAN by maximizing $\mathcal{L}_{D2}$ and minimizing $\mathcal{L}_{G2}$ given in equations \ref{eq:stage2d} and \ref{eq:stage2g}.

\begin{equation} \label{eq:stage2d}
\begin{aligned}
\mathcal{L}_{D2} = \mathbb{E}_{(I_2, \psi(t)) \thicksim p_{data}}[\log D_2(I_2, \psi(t)] +\\
\mathbb{E}_{I_2 \thicksim G_2, \psi(t) \thicksim p_{data}}[\log (1 - D_2(I_2, \psi(t))]
\end{aligned}
\end{equation} 
\begin{equation} \label{eq:stage2g}
\begin{aligned}
\mathcal{L}_{G2} = \mathbb{E}_{z \thicksim \mathcal{N}(0,1), \psi(t) \thicksim p_{data}}[\log (1 - D_2(G_2(z, \psi(t)))]
\end{aligned}
\end{equation}

But unlike Stage-1 GAN, we do not introduce random noise $z$.

\subsubsection{Total Image Synthesis GAN Loss}

We can combine all the objective functions for the image synthesis network given in the preceding sections into one objective function $\mathcal{L}_{imageGAN}$ given in equation  \ref{eq:imageganloss}.
\begin{equation}
\begin{aligned} \label{eq:imageganloss}
\mathcal{L}_{imageGAN}(G, D) = \mathcal{L}_{G1-Total} + \mathcal{L}_{D1} + \mathcal{L}_{G2} + \mathcal{L}_{D2}
\end{aligned}
\end{equation}
and we aim to solve:
\begin{equation} \label{eq:solveforward}
\begin{aligned}
G1^*, G2^*, D1^*, D2^* = \\
\argmin_{G1, G2} \argmax_{D1, D2} \mathcal{L}_{imageGAN}(G, D)
\end{aligned}
\end{equation}
\subsection{Image to Text Translation}
We then built an image captioning network that produces captions based on images generated from the image synthesizing GAN. The generated captions can reinforce cycle consistency. Our work is based on Dai et al. \cite{DBLP:journals/corr/DaiLUF17}.
\subsubsection{Architecture}
The architecture is shown in figure 3. The generator $F$ takes an image $I$ concatenated with noise $z$ as input to an LSTM. The LSTM then produces a caption $S$. The generated caption $S$ is then evaluated by the discriminator $E$, where it outputs a matching reward $r$. 

\subsubsection{Overall loss}
According to the GAN formulation, we can derive the overall loss function as:

\begin{equation} \label{eq:captiongan1}
\begin{aligned}
\min_{\theta} \max_{\eta} \mathbb{E}_{S \thicksim P_I}[\log r_{\eta}(I,S)] + \\
\mathbb{E}_{z \thicksim N_0}[\log (1-r_{\eta}(I,F_{\theta}(I,z)))]
\end{aligned}
\end{equation}

where $\theta$ is the parameters of the caption generator $F$ and $\eta$ is the parameters of the caption discriminator $E$. $S$ is the sampled caption from the caption generator. $r$ is the reward calculated by the discriminator using equation \ref{eq:captiongan2}.

\begin{equation} \label{eq:captiongan2}
\begin{aligned}
r_{\eta}(I,S) = \sigma(\langle f(I, \eta_I), h(S, \eta_S) \rangle)
\end{aligned}
\end{equation}

where $\sigma$ is the sigmoid function, $f(.)$ is the feature vector obtained by convolutional layers, $h(.)$ is the embedding produced by the LSTM, and $\langle,\rangle$ denotes dot-product operation.
\subsubsection{Discriminator Training}
The discriminator loss consists of 3 terms: 1. Real loss, which enforces the discriminator to give high score for real image with real caption $S_I$. 2. Fake loss, which makes sure that the discriminator can differentiate fake captions $S_G$ generated from the generator $F$. 3. Wrong loss, which makes sure that the discriminator does not associate wrong caption $S_{\backslash I}$ with a real image. So the total loss is:

\begin{equation} \label{eq:captiongan2}
\begin{aligned}
\mathbb{E}_{S \in S_I}[\log r_{\eta}(I,S)] + \alpha \cdot \mathbb{E}_{S \in S_G}[\log (1-r_{\eta}(I,S))] + \\
\beta \cdot \mathbb{E}_{S \in S_{\backslash I}}[\log (1-r_{\eta}(I,S))]
\end{aligned}
\end{equation}

\subsubsection{Generator Training}
We are training the generator similar to Dai et al. \cite{DBLP:journals/corr/DaiLUF17}. However, because of the computation limitations, we slightly modified the loss function. To be more specific, instead of taking the expected value of the reward by running Monte Carlo estimate, we just take one sample as the expectation. Furthermore, in order to decrease the computational complexity, we only perform Monte Carlo rollout for the most likely word $w$ instead for all possible words. Therefore, our final generator loss is:

\begin{equation} \label{eq:captiongan2}
\begin{aligned}
\mathbb{E}[\sum^{T_{max}}_{t=1} \argmax_{w_t \in V} \pi_{\theta}(w_t|I, S_{1:t-1}) \cdot r_{\eta}(I,S)]
\end{aligned}
\end{equation}

$S$ in the above equation has 3 parts, the first part is $S_{1:t-1}$ taken from the real caption, the second part is $S_t$ which is the chosen word from the argmax function, the third part is $S_{t+1:T}$ sampled from the Monte Carlo rollout. This alternative interpretation allows us to decrease the computational complexity by a factor of $V \cdot T$ where $V$ is the vocabulary size and $T$ is the maximum time steps. 

\subsection{Cycle Consistency}

The generator $G$ ($G1$ and $G2$ combined) of image synthesizing network learns a domain mapping from text to images, while the generator of the image captioning network $F$ learns an inverse mapping, from images back to text.

To improve the results of the image synthesizing network further, we can exploit using the law of transitivity by introducing cycle consistency as showed in \cite{zhu2017unpaired}\cite{liu2017unsupervised}\cite{zhou2016learning}.

Zhu et al. \cite{zhu2017unpaired} use cycle consistency by introducing two additional loss terms, forward cycle loss and backward cycle loss. The main reason to use these additional loss terms is because learning a mapping from one image domain to another without a paired dataset is an under-constrained problem.

The additional loss terms penalize the network parameters if they cannot reconstruct the original image by using the law of transitivity, i.e $F(G(x))\approx x$ and $G(F(y))\approx y$.

Since we are primarily interested in improving the results of the image synthesizing network, we only use forward cycle loss to reconstruct the original caption back from the generated image. Hence we define forward cycle loss as the cross entropy loss between generated word and the actual word in the training dataset. We represent forward cycle loss as $\mathcal{L}_{fcycle}$ and is defined in the equation \ref{eq:cyclecrossentropy}.

\begin{equation} \label{eq:cyclecrossentropy}
\begin{aligned}
\mathcal{L}_{fcycle} = - \sum_{t=0}^{T-1} \log p_t(w_t|I)
\end{aligned}
\end{equation}

Here $T$ is the number of words in original caption from the train dataset. $p_t(w_t|I)$ is the probability of observing the correct word $w_t$ generated by the LSTM in the captioning network given an image $I = G(z, \psi(t))$ generated by the image synthesis network's generator.

We can now express the objective function for training all the networks as shown in the equation \ref{eq:allfinal}.
\begin{equation}\label{eq:allfinal}
\mathcal{L}_{final} = \mathcal{L}_{imageGAN}(G, D) + \mathcal{L}_{textGAN}(F,E) + \lambda_{c} \mathcal{L}_{fcycle}
\end{equation}

Where $\lambda_c$ is hyperparameter that decides the importance of forward cycle loss. For our experimentation, we use $\lambda_c = 2$.

\begin{figure*}[t]
\begin{center}
	\includegraphics[width=\textwidth]{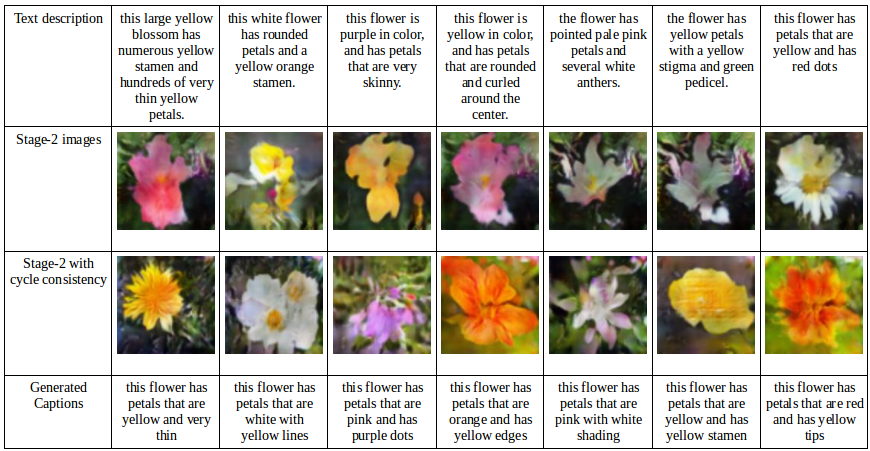}
    \caption{First row contains ground truth text descriptions, next two rows contain images generated by the GAN trained without cycle loss and trained with cycle loss respectively. The last row contains captions generated by the captioning network}
    \label{fig:cycle}
\end{center}
\end{figure*}
\begin{figure*}[t] 
\begin{center}
	\includegraphics[width=\textwidth]{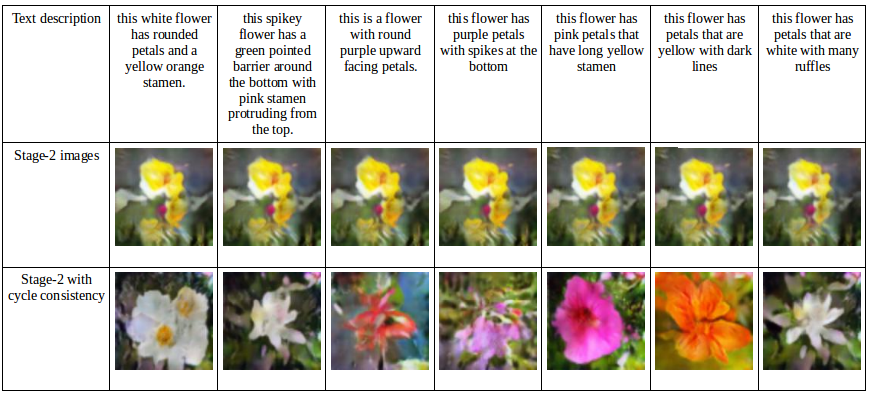}
    \caption{Figure illustrating mode collapse. The first row contains ground truth text descriptions. The next two rows contain images generated by the GAN trained without cycle loss and with cycle loss respectively.}
    \label{fig:modecollapse}
\end{center}
\end{figure*}
\section{Dataset}

We primarily use the Oxford VGG 102 Flower Dataset \footnote{http://www.robots.ox.ac.uk/~vgg/data/flowers/102/index.html} \cite{Nilsback08} for our experiments. This dataset contains 102 categories of flowers. Each category contains images of flowers between 40 and 248, with 8,189 images in total.

For the flowers dataset, we use 5 captions per image. This gives us 5 image, caption pairs for every image present in the dataset. The images provided don't have a fixed dimension. We resize the images to 64x64 and 128x128 resolutions to be used by our network.

We generate Skip-Thought text embeddings of all captions before the experimentation, and construct image, embedding pairs. Hence we have five data points for every image that is present in the dataset.
\section{Experiments}

We first pretrain the image synthesis GAN network for 100 epochs, and pretrain the image captioning GAN network for 100 epochs. We then combine the two and train the whole network end to end by optimizing the objective $\mathcal{L}_{total}$ which includes the cycle loss for another 40 epochs.

Throughout, we use a base learning rate of 0.0001, ADAM optimizer with $\beta_1 = 0.5$ and $\beta_2 = 0.999$. We use  mini-batch size of 64. For text features, we use a Skip-Thought network to generate embeddings which are 2400-dimensional vectors. Our noise vector is a 100-dimensional standard gaussian distribution.

All of our code is available at: \url{https://github.com/CSC2548/text2image2textGAN}

\section{Results}
\subsection{Text to Image Translation Results}
We first present Stage 1 GAN results for VGG 102 Flowers Dataset and compare the quality of 64x64 generated images with Stage 2 GAN results.

The figure \ref{fig:stage1} refers to the results produced by Stage 1 GAN. The first row contains the ground truth text descriptions of the image. The last row contains the images generated using our Stage 1 GAN.

As we can see the GAN learns many interesting features such as color of the flower, size of the flower, etc. We can observe that the GAN even learned a few subtle features such as "spiky petal".

Stage 2 GAN produces higher resolution images refining the output produced by Stage 1 GAN. The results are shown in the figure \ref{fig:stage2}.

\subsection{Image to Text Translation Results}

Our text captioning GAN produces captions based on input images. The results are shown in the figure FILL. The first row contains ground truth images from the dataset, the following row shows the captions generated for these images.

\subsection{Cycle Consistency Results}

We train our image synthesis and image captioning networks end to end after pretraining them individually. We observed many improvements using cycle consistency.

Observing results produced in figure \ref{fig:cycle}, we can see that the GAN produces results that are different from the text descriptions. For example the caption "This flower has yellow petals with red dots" did not generate such a flower, but rather generated a white flower, with a yellow center. Training with cycle loss makes the network update the parameters of image synthesizing network so that the captioning network produces accurate captions close to the ground truth captions. This helps the network generate more accurate images that describes the text well.

We also observe significant reduction in mode collapse. Figure \ref{fig:modecollapse} shows images generated before training with cycle loss and after training with cycle loss. We can observe that the network without the cycle loss produces similar images, but network trained with cycle loss has more diverse images that are generated.

\begin{figure}[H]
\begin{center}
	\includegraphics[height=8cm,width=14cm,keepaspectratio]{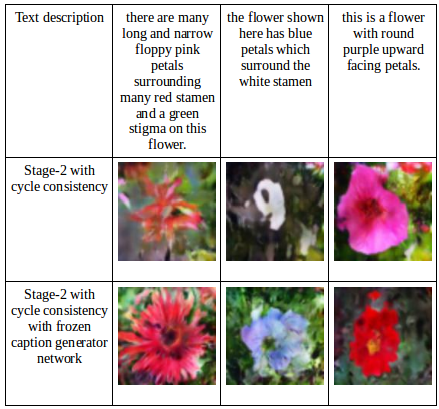}
    \caption{Results of GAN trained with cycle loss with image captioning network's weights frozen}
    \label{frozen}
\end{center}
\end{figure}

Since training with cycle loss could also sometimes introduce bias in the captioning network as this would force the network to produce captions close to ground truth captions on noisy and wrongly generated images, we froze the weights of captioning network and show a few sample images in figure \ref{frozen}.

\begin{table}
\begin{tabular}{ |p{2.7cm}||p{2cm}|p{2cm}|  }
 \hline
 \multicolumn{3}{|c|}{Inception Score} \\
 \hline
 Model & Mean & Standard Deviation\\
 \hline
 \hline
 GAN without Cycle Loss & 2.985 & 0.163\\
 \hline
 GAN with Cycle Loss   &  2.545 & 0.067\\
 \hline
\end{tabular}
\caption{Inceptions scores comparing GAN trained with cycle loss and trained without cycle loss}
\label{table:inception}
\end{table}

\subsection{Inception Score}

We tabulate the inception score comparing our text to image GAN trained with cycle consistency and without cycle consistency for over 2500 random images. However the effectiveness is debatable as shown in \cite{barratt2018note}.

\subsection{Image color relevance score}

To analyze the effectiveness of our method further, we manually find the ratio of number of colors that are present in each image to all the colors that were mentioned in the caption. We average this number over 30 random images generated by our GAN model trained with cycle consistency and trained without cycle consistency. We call this as color relevance score and show the results in table \ref{table:colorrelevance}. We observe significant improvement in our model's ability to generate the images with colors as described in the caption.

\begin{table}
\begin{tabular}{ |p{4cm}||p{3.09cm}|  }
 \hline
 Model & Color relevance score\\
 \hline
 \hline
 GAN without Cycle Loss & 0.259 \\
 \hline
 GAN with Cycle Loss   &   0.802\\
 \hline
\end{tabular}
\caption{Image color relevance score for the two models}
\label{table:colorrelevance}
\end{table}

\section{Conclusion}

In this work, we implement a text-to-image translation GAN and image-to-text translation GAN, using existing popular methods. We show that these methods sometimes produce non-relevant images based on the text description given as input to the model and also suffers from mode-collapse. We improve the results by enforcing cycle consistency by generating captions on the generated images and further optimizing the network to reduce the distance between the generated text and ground truth text. In future, we aim to generate higher quality images and test on complicated datasets such as MS-COCO. We also think think there is good potential in comparing the semantic meaning of text using fixed length embeddings measuring the distance in cycle loss.
\bibliographystyle{plain}
\bibliography{ref}

\end{document}